\documentclass[conference]{IEEEtran}
\ifCLASSINFOpdf
\else
\fi
\usepackage{algorithm}
\usepackage{algorithmic}

\usepackage[utf8]{inputenc} % allow utf-8 input
\usepackage[T1]{fontenc}    % use 8-bit T1 fonts
\usepackage{hyperref}       % hyperlinks
\usepackage{url}            % simple URL typesetting
\usepackage{booktabs}       % professional-quality tables
\usepackage{amsfonts}       % blackboard math symbols
\usepackage{nicefrac}       % compact symbols for 1/2, etc.
\usepackage{microtype}      % microtypography
\usepackage{xcolor}         % colors

\usepackage{inconsolata}
\usepackage{paralist}
\usepackage{amsmath}
\usepackage{amsthm}
\usepackage{booktabs}
\usepackage{algorithm}
\usepackage{algorithmic}
\usepackage{subfigure}
\usepackage{amssymb}
\usepackage{graphicx}
\usepackage{subfigure}
\usepackage{multirow}

\begin{document}
\title{Enhancing Topic Interpretability for Neural Topic Modeling through Topic-wise Contrastive Learning}

% \author{\IEEEauthorblockN{Michael Shell}
% \IEEEauthorblockA{School of Electrical and\\Computer Engineering\\
% Georgia Institute of Technology\\
% Atlanta, Georgia 30332--0250\\
% Email: http://www.michaelshell.org/contact.html}
% \and
% \IEEEauthorblockN{Homer Simpson}
% \IEEEauthorblockA{Twentieth Century Fox\\
% Email: homer@thesimpsons.com}
% \and
% \IEEEauthorblockN{James Kirk\\ and Montgomery Scott}
% \IEEEauthorblockA{Starfleet Academy\\
% San Francisco, California 96678--2391\\
% Telephone: (800) 555--1212\\


\author{\IEEEauthorblockN{Xin Gao\IEEEauthorrefmark{2}\IEEEauthorrefmark{4},
Yang Lin\IEEEauthorrefmark{2}\IEEEauthorrefmark{4},
Ruiqing Li\IEEEauthorrefmark{2}\IEEEauthorrefmark{4}, 
Yasha Wang\IEEEauthorrefmark{1}\IEEEauthorrefmark{2}\IEEEauthorrefmark{5},
Xu Chu\IEEEauthorrefmark{3},
Xinyu Ma\IEEEauthorrefmark{2}\IEEEauthorrefmark{4} and
Hailong Yu \IEEEauthorrefmark{2}\IEEEauthorrefmark{4}\\
Email: xingao@pku.edu.cn, bdly@pku.edu.cn, lrq@stu.pku.edu.cn, wangyasha@pku.edu.cn,\\chu\_xu@pku.edu.cn, maxinyu@pku.edu.cn, and hailong.yu@pku.edu.cn
}
\IEEEauthorblockA{\IEEEauthorrefmark{4}Software Engineering Institution, School of Computer Science,  Peking University, Beijing, China}
\IEEEauthorblockA{\IEEEauthorrefmark{2}Key Lab of High Confidence Software Technologies, Ministry of Education, Beijing China\\  } \IEEEauthorblockA{\IEEEauthorrefmark{3}Center on Frontiers of Computing Studies, School of Computer Science, Peking University\\ }
\IEEEauthorblockA{\IEEEauthorrefmark{5}National Engineering Research Center of Software Engineering, Peking University, Beijing, China}
\IEEEauthorblockA{\IEEEauthorrefmark{1}Corresponding author}
}


\maketitle

\begin{abstract}
Data mining and knowledge discovery are essential aspects of extracting valuable insights from vast datasets. Neural topic models (NTMs) have emerged as a valuable unsupervised tool in this field. However, the predominant objective in NTMs, which aims to discover topics maximizing data likelihood, often lacks alignment with the central goals of data mining and knowledge discovery which is to reveal interpretable insights from large data repositories. Overemphasizing likelihood maximization without incorporating topic regularization can lead to an overly expansive latent space for topic modeling.
In this paper, we present an innovative approach to NTMs that addresses this misalignment by introducing contrastive learning measures to assess topic interpretability. We propose a novel NTM framework, named ContraTopic, that integrates a differentiable regularizer capable of evaluating multiple facets of topic interpretability throughout the training process. Our regularizer adopts a unique topic-wise contrastive methodology, fostering both internal coherence within topics and clear external distinctions among them.
Comprehensive experiments conducted on three diverse datasets demonstrate that our approach consistently produces topics with superior interpretability compared to state-of-the-art NTMs. 
% Supplementary materials, including code and the 20NG dataset used in our study, are available for reference\footnote{https://anonymous.4open.science/r/ContraTopic-CACD}.
\end{abstract}

\section{Introduction}

%第一段
% As an unsupervised statistical tool, topic modeling aims at discovering a set of latent semantic topics from a collection of text documents.
% A popular topic model is the Latent Dirichlet Allocation (LDA)~\cite{blei2003latent}, where the authors construct the generative story based on the Dirichlet prior and develop a variational Bayesian algorithm to perform approximate inference.

% Topic modeling, serving as an unsupervised statistical tool, strives to uncover latent semantic topics within a corpus of textual documents, with the Latent Dirichlet Allocation (LDA)~\cite{blei2003latent} being a well-known and widely used model that utilizes a generative story based on the Dirichlet prior and employs a variational Bayesian algorithm for approximate inference.
% Topic interpretability is crucial in topic modeling as it allows researchers and users to gain meaningful insights, understand underlying patterns, and effectively utilize the extracted topics for various applications~\cite{zhang2017survival,wood2017source, hoyle2021automated, jelodar2019latent,ziyue2021tensor,wang2022intent}.

In the realm of data mining and knowledge discovery, topic modeling stands as a fundamental unsupervised statistical technique. Its primary aim is to unveil latent semantic topics within a corpus of textual documents. One prominent model in this domain is Latent Dirichlet Allocation (LDA)\cite{blei2003latent}. LDA relies on a generative framework grounded in the Dirichlet prior and employs variational Bayesian algorithms for approximate inference. Crucially, topic interpretability plays a pivotal role in the context of data mining, as it empowers researchers and users to extract meaningful insights, comprehend underlying patterns, and effectively employ the identified topics for diverse applications\cite{zhang2017survival,wood2017source, hoyle2021automated, jelodar2019latent,ziyue2021tensor,wang2022intent}.

%The topic interpretability of traditional topic models such as Latent Dirichlet Allocation (LDA)~\cite{blei2003latent} is usually implicitly ensured by their Dirichlet prior which encourages sparsity in topic-word distribution.

Recent years have borne witness to an unprecedented surge in the scale of corpora requiring analysis through topic modeling. Nonetheless, generating high-quality topics on substantial corpora using LDA and its extensions has been recognized as a formidable challenge~\cite{miao2016neural,srivastava2017autoencoding}. To address these challenges and enhance adaptability and scalability, researchers have integrated Auto-encoding Variational Bayes~\cite{kingma2013auto} into the inference process. This integration has given rise to a class of models known as neural topic models (NTMs)\cite{ miao2016neural, srivastava2017autoencoding, dieng2020topic}. However, despite these advancements, some NTMs have faced criticism for generating topics that are less interpretable to humans\cite{zhao2020neural}.

%Nevertheless, devising suitable reparameterization tricks for the Dirichlet distribution poses a challenging task, leading researchers to employ diverse approximations in NTMs, which ultimately compromises the interpretability of their generated topics.

The main obstacle hindering NTMs from generating highly interpretable topics stems from the inconsistency between the current objective, \textit{which focuses on discovering topics that maximize the likelihood of observed data} and the users' intention in topic modeling, \textit{which includes discovering topics with high interpretability that facilitate human comprehension of large corpora}.
By solely focusing on likelihood maximization without incorporating any regularization, the latent space for topic modeling may become excessively large to comprehensively explore.
Traditional topic models like LDA implicitly maintain topic interpretability by leveraging their Dirichlet prior, which promotes sparsity in topic-word distribution to restrict the latent space.
Nevertheless, devising appropriate and accurate reparameterization tricks for the Dirichlet distribution poses a challenging task, leading researchers to employ various approximations in NTMs, which ultimately compromises the topics' interpretability.
Consequently, we propose to \textbf{integrate measurements of topic interpretability into the objective to regulate the latent space and steer NTMs toward uncovering better topics.}

At present, the evaluation of topic interpretability typically revolves around two aspects: \textit{topic coherence} and \textit{topic diversity}, where coherence measures the degree of semantic consistency among the most relevant words within a topic, while diversity gauges the distinctiveness of the most relevant words across different topics as shown in Figure~\ref{fig:contrastive}.
% \textit{Topic coherence} measures the coherence among the most related words from the same topic while \textit{topic diversity} measures the uniqueness among the most related words from different topics.
In order to constrain the latent space, we propose an idea that \textbf{explicitly incorporates topic interpretability as a regularizer into the objective function of NTMs}. 
While the idea appears intuitive, several challenges persist that require further investigation.

\textbf{Challenge 1: Developing a computationally friendly regularizer.}
One approach is to directly incorporate interpretability-based evaluation metrics as regularizers.
However, these metrics, e.g., \textit{topic coherence}, rely on look-up operations in a large reference corpus, making them non-differentiable and computationally intensive~\cite{ding2018coherence}.
Such non-differential operations may make it challenging to calculate gradients and perform backpropagation.
Even though methods like policy gradient can address the non-differentiability issue, they usually introduce higher computational complexity and gradient variance.
A preferable solution would involve designing differentiable surrogates that capture human cognition and encompass various aspects of topic interpretability.

\textbf{Challenge 2: Addressing the consideration of multiple aspects in topic interpretability?}
Drawing from relevant literature~\cite{doogan2021topic,dieng2020topic, nan2019topic, chang2009reading, meng2020discriminative}, it is widely recognized that topic interpretability necessitates the generation of topics that are both semantically coherent and distinctive to human understanding.
Neglecting the aspect of distinctiveness may result in redundant topics with overlapping global themes, while overlooking coherence can result in superficially distinctive topics lacking a coherent semantic meaning.
% Without consideration of distinction, there will be redundant topics with similar global themes.
% Without consideration of coherence, the topics will be somehow distinctive, but their most related words will have few connections and be difficult to understand.
Balancing coherence and distinctiveness among topics remains a challenging task.
To address this, we propose a topic-wise contrastive term that different from previous document-wise contrastive terms in NTMs.
The core concept of contrastive learning involves promoting similarity between positive sample pairs while discouraging similarity between negative sample pairs.
In the context of Contratopic, positive samples consist of words sampled from the same topic, while negative samples encompass words from different topics.
Therefore, encouraging positive samples promotes coherence within a topic, while discouraging negative samples fosters distinctiveness across topics.

Within this paper, we propose an innovative framework for NTMs that introduces topic interpretability during the training phase through the integration of a topic-wise contrastive learning regularizer.
This regularizer treats words associated with the same topic as positive samples and words associated with different topics as negative samples, thus enhancing the coherence and distinctiveness of the topic-word distribution.
% %From perspective of information theory, our topic-wise contrastive regularizer simultaneously estimates and maximizes the mutual information between the word distribution of different topics.
We additionally utilize a novel sampling technique that leverages the Gumbel-Softmax trick~\cite{jang2016categorical}, enabling us to efficiently draw top-k related words from the topic-word distribution in a differentiable manner, generating pairs of positive and negative samples without replacement.
To evaluate the effectiveness of our regularizer in improving topic interpretability, we conducted extensive experiments on three datasets.
The results demonstrate the superior performance of our proposed ContraTopic approach compared to all baseline methods.

The contribution can be summarized as follows:
\begin{itemize}
    \item To mitigate the inconsistency between the current objective of NTMs and the intent usage of topic modeling, we propose an innovative framework for NTMs, named ContraTopic.
    \item Based on the multiple aspects of topic interpretability, we devise a differentiable regularizer via topic-wise contrastive learning, promoting coherence within individual topics and diversity across different topics. 
    \item  Through experiments conducted on three datasets, we demonstrate that ContraTopic consistently generates topics with significantly higher levels of interpretability compared to all baseline approaches.    
\end{itemize}

\section{Related Works}
In this section, we will first briefly introduce the development and main branches of the NTMs based on their architectures.
Then we focus on NTMs that also leverage contrastive learning but with different goals and methodologies.
Last but not least, we introduce NTMs that also attempt to improve the interpretability of generated topics.
\subsection{Neural Topic Models}
A major challenge in applying conventional topic models and developing new methods was the complex inference algorithms of the posterior distribution.
The employment of neural networks and black-box inference methods has enabled topic models to automatically solve the inference process via backpropagation, which led to the thriving of NTMS.
NVDM ~\cite{miao2016neural} and ProdLDA ~\cite{srivastava2017autoencoding} leveraged autoencoding variational Bayes~\cite{kingma2013auto} as the basic architecture and respectively applied Gaussian and logistic normal distribution in approximations of the Dirichlet prior in the original LDA.
Subsequently, various constructions of the prior distributions have been proposed ~\cite{zhang2018whai,nan2019topic,burkhardt2019decoupling,dieng2020topic} aiming for a better approximation of the Dirichlet prior.
On top of variational autoencoders (VAE), other famous architectures, such as generative adversarial networks~\cite{wang2019atm,wang2020neural} and graph neural networks~\cite{yang2020graph, xie2021graph}, have also been applied into NTMs.
Recently, many researchers have proposed novel NTMs via the theory of optimal transport (OT).
By naturally incorporating word embeddings into the cost function of the optimal transport distance, their models~\cite{zhao2020neural, wang2022representing} were able to achieve a better balance between obtaining good document representation and generating topics with high quality.
Recently, ECRTM~\cite{wu2023effective} proposes a novel embedding clustering regularization that avoids the collapsing of topic embeddings.

\subsection{Contrastive Learning in NTMs}
The idea of contrastive learning is to measure the similarity relations between samples by contrasting positive pairs against negative pairs.
There have been various efforts to study contrastive methods to learn meaningful representation in a self-supervised way.
It has been widely explored in enormous fields including image classification~\cite{hjelm2018learning,khosla2020supervised}, object detection~\cite{xie2021detco,sun2021fsce}, adversarial training~\cite{miyato2018virtual}, and sequence modeling~\cite{logeswaran2018efficient}. 
In recent years, contrastive learning also has been applied in topic modeling to leverage the relations among documents~\cite{zhou2023improving}.
Nguyen et al.~\cite{nguyen2021contrastive} proposed CLNTM with a novel contrastive objective that captured the mutual information between the document prototypes and their positive samples by modeling the relations among augmented samples.
TSCTM~\cite{wu2022mitigating} was proposed to sufficiently model the relations between documents against data sparsity via a new contrastive learning method with efficient sampling strategies.

\textbf{A pivotal distinction between our method and other topic models that incorporate contrastive terms} stems from the foundational insights underlying each approach. Other methods, such as CLNTM, employ a \textbf{document-wise} contrastive term, with their guiding principle centered around \textbf{encouraging similar document-topic distribution among similar documents}, which will "implicitly" benefit the topic-word distribution and thereby contribute to topics with higher quality. In stark contrast, our approach adopts a \textbf{topic-wise} contrastive term, which, in a more direct and unequivocal manner, \textbf{fosters coherence and diversity of the topic-word distribution}, leading to an enhancement in the quality of topics. Empirical experiments corroborate this distinction, revealing that our topic-wise contrastive term eclipses the quality achieved by CLNTM, thus underscoring the  merits of directly optimizing the topic-word distribution.
More details about the differences can be found in subsection G of section Methodology.
\subsection{Improving Topic Interpretability in NTMs}
%Since topic models were proposed to assist humans in understanding the latent semantic topics of large corpora, most of the works aimed at improving the interpretability of generated topics.
% For evaluation, lots of efforts~\cite{chang2009reading,doogan2021topic,hoyle2021automated} have been made to design metrics that are closer to human cognition so that models that generate better topics can be selected.
There have been lots of works aiming at improving the topic interpretability of NTMs.
Some works~\cite{miao2016neural,srivastava2017autoencoding,zhang2018whai,nan2019topic} focused on finding better approximations for Dirichlet distribution, which encourages the sparsity among topics and implicitly improves the topic interpretability.
%Besides, the success of LDA gives the main credit to the Dirichlet prior, which encourages the sparsity among topics and implicitly improves the topic interpretability.
%However, due to the intractability of the Dirichlet distribution, plenty of NTMs~\cite{miao2016neural,srivastava2017autoencoding,zhang2018whai,nan2019topic} have to resort to various approximations of the Dirichlet distribution to the compromise.
%Another popular way to improve topic interpretability is to incorporate other meta information into the modeling process, such as seed words~\cite{harandizadeh2022keyword} and document labels~\cite{wang2020neural}, that are closely related to topics.
Besides, another popular way to improve topic interpretability is to incorporate other external meta information into topic modeling, such as seed words~\cite{harandizadeh2022keyword,lin2023enhancing} and document labels~\cite{wang2020neural}, that are closely related to topics.
Different from these methods, we do not incorporate such external information, which means that there are no extra human-annotation costs and expertise in ContraTopic.
Apart from incorporating external information, an alternative straightforward solution is to incorporate topic interpretability on the specific corpus into the objectives.
Gui et al.~\cite{gui2019neural} borrow the idea of reinforcement learning and incorporate topic coherence measures as rewards to guide the NTMs.
However, their approach updates the topic-word distribution using rigid pre-defined rules, and the intricate complexity of the states poses challenges for achieving convergence.
Ding et al.~\cite{ding2018coherence} also propose a method to incorporate topic coherence based on word embeddings that is differentiable and computation-efficient, but their method only focuses on topic coherence and ignores other aspects of topic interpretability such as \textit{topic diversity}.
In contrast, our approach uses the pre-computed NPMI in the given corpus to provide extra supervision and introduces a topic-wise contrastive regularizer that incorporates multiple facets of topic interpretability.

\section{Preliminary}
\subsection{Problem Formulation}
Consider a corpus consisting of $D$ documents, where the vocabulary contains $V$ distinct terms.
Each document is represented as a bag-of-word vector $x\in \mathcal{R}^V$.
Our goal is to derive $K$ topics, which are set of distributions over words noted as $\beta_{1},...,\beta_{K} \in \Delta^{V}$, and latent topic distribution for each document noted as $\theta_{1},...\theta_{D} \in \Delta^K$, where $\Delta^{K}$ is a $K-1$ dimensional simplex.
\subsection{Neural Topic Modeling}
The common generative story of NTMs is similar to \cite{srivastava2017autoencoding}, where the Dirichlet prior is approximated via a logistic normal distribution.
The generative story is summarized as follows, where $\alpha$ is the parameter for prior distribution:
\begin{compactenum}
    \item[] 1. Draw topic distribution $\theta\sim\mathcal{LN}(\mu_0(\alpha), \sigma_0^2(\alpha))$;
        \item[] 2. For $w_{dn}$ in this document:
        \begin{compactenum}
            \item[] a. Draw topic $z_{dn}\sim Cat(\theta)$;
            \item[] b. Draw word $w_{dn}\sim Cat(\beta_{z_{dn}})$;
        \end{compactenum}
\end{compactenum}
The subscripts like $w_{dn}$ on the third page indicate the n-th word in the d-th document of the corpus, following conventions from seminal topic model papers~\cite{blei2003latent,dieng2020topic}.
The $\mathcal{LN(\cdot)}$ denotes the logistic-normal distribution, and $Cat(\cdot)$ denotes the multinomial distribution.
To leverage the pre-trained word embeddings, ETM~\cite{dieng2020topic} uses the  word embeddings for the vocabulary $\mathbf{\rho} \in \mathcal{R}^{V\times e}$, where $e$ is the dimension of embedding, and the assigned topic embedding $t_{z_{dn}} \in \mathcal{R}^e$ to draw the observed word from the assigned topic $z_{dn}$, noted by $\beta_{z_{dn}} = \mathrm{softmax}(\mathbf{\rho} t_{z_{dn}}/ \tau_{\beta})$, where $\tau_{\beta}$ is a temperature forcing the distribution to be sharp.

Based on the generative story, the variational inference is used to approximate the posterior distribution of latent variables $\rho$, $\theta_d$, $\mathbf{z}_{d}=\{z_{d1},z_{d2},\dots,z_{dN_d}\}$ and $\mathbf{t}=\{t_{1},t_{2},\dots,t_{K}\}$ to maximize the likelihood of observed data. 
The evidence lower bound (ELBO) can be derived as 
\begin{equation}\label{eq:elbo}
\small
\begin{aligned}
    \mathcal{L}(\boldsymbol{w}) = &E_{q(\theta,\boldsymbol{z}|\boldsymbol{w})}\log \left(p(\boldsymbol{w}|\theta,\boldsymbol{z};\boldsymbol{t})\right) 
    - E_{q(\theta,\boldsymbol{z}|\boldsymbol{w})}\log \left(\frac{q(\theta,\boldsymbol{z}|\boldsymbol{w})}{p(\theta, \boldsymbol{z})}\right).
\end{aligned}
\end{equation}
The first term noted as $\mathcal{L}_{rec}$ encourages variational distribution $q$ to favor information that is good at explaining the observed words, and the second term noted as $\mathcal{L}_{kl}$ encourages $q$ to match the prior distribution.

Specifically, for the encoder $q(\theta|\boldsymbol{w})$, we have $\pi=\mathrm{}{MLP}(w)$, $\mu(w) =l_1(\pi)$, $log \sigma(w)=l_2(\pi)$ and finally $\theta(w)=\mathrm{softmax} (\mu+\sigma \odot \epsilon)$, where $\epsilon \sim \mathcal{N}(0,I)$. The functions $l_1$ and $l_2$ are linear transformations.
For the decoder network, we collapse $\boldsymbol{z}$ and compute $p(w|\theta,\rho,\boldsymbol{t})=\theta^\top\boldsymbol{\beta}  $.

\section{Methodology}
\subsection{Topic-wise Contrastive Regularizer}
\begin{figure*}[h]
\centering
  \includegraphics[width=6 in]{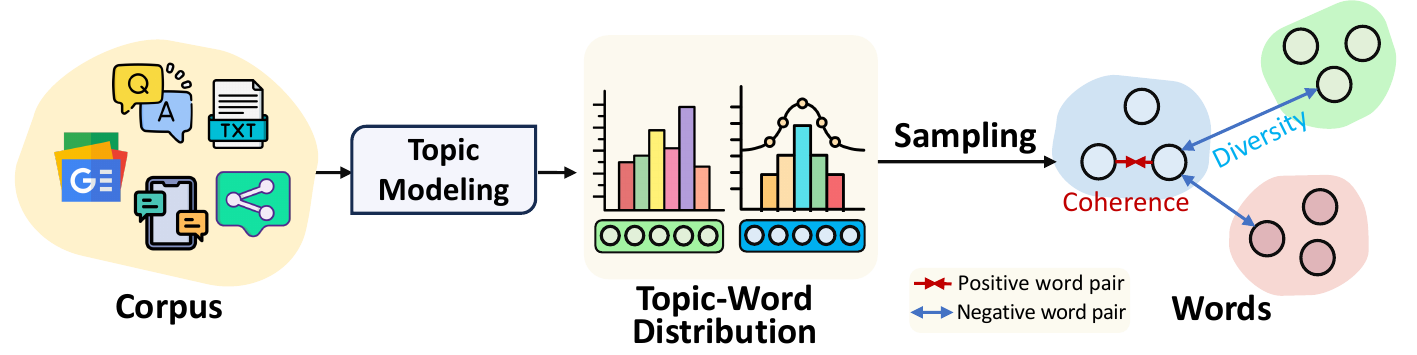}
  \caption{The fundamental insight of ContraTopic. During the training of NTMs, words are sampled from each topic for the evaluation of topic coherence and diversity. Words in the same color are sampled from the same topic. By encouraging similarity between positive word pairs and discouraging similarity between negative word pairs, the  coherence and diversity of generated topics can be improved.   }
  \label{fig:contrastive}
\end{figure*}

As shown in the ELBO, the current objective of NTMs lacks consideration of topic interpretability during the training process of VAE-based NTMs, making it difficult to control the topic quality.
To incorporate topic interpretability, we propose a topic-wise contrastive regularizer that simultaneously controls the coherence and distinction of generated topics. Figure~\ref{fig:contrastive} shows the fundamental insight of ContraTopic.

Following~\cite{khosla2020supervised}, we first introduce the contrastive loss with multiple positive samples and negative samples\footnote{It should be noted that the samples in our contrastive regularizer refer to words, not documents.}.
For a set of M samples $\boldsymbol{S}={s_1,....s_M}$, the contrastive loss can be calculated as Eq.\ref{eq:contrastive}.
\begin{equation}\label{eq:contrastive}
\small
\mathcal{L}_{con}=\sum_{i \in I} {-}\log \left( \frac{\sum_{p \in P(i)}\exp \left(\mathcal{K}(i , p) \right)}{\sum_{a \in I} \exp \left(\mathcal{K}(i, a)\right)} \right),
\end{equation}
where $I \equiv\{1 \dots M\}$ is the full set of the indices of all samples, $P(i)$ is the set of indices of all positive samples for anchor i.

Concretely, the full set of samples $\mathcal{S}$ consists of $K$ sets of words that are sampled from each topic, noted by $\mathcal{S}= \bigcup_{k=1}^K T_i$, where $T_k$ denotes the most $v$ related words in the $k$-th topic, $v=\frac{M}  {K}$.
$P(i)$ denotes the indices of samples from the same topic as $s_i$, noted by $P(i)= \{j|s_i \in T_k, s_j \in T_k$,  and $ i \neq j \}$.
The function $\mathcal{K}(\cdot)$ measures the similarity between two words.
It can be implemented with dot product of word embeddings or the pre-computed Normalized Point-wise Mutual Information (NPMI) in the corpus.
By employing the pre-computed NPMI in the corpus for $\mathcal{K}(\cdot)$, the positive samples in our regularizer serve to optimize topic coherence directly during training, while the negative samples aim to encourage the sampling of words with low NPMI from distinct topics. 

\textbf{The choice of NPMI in $\mathcal{K(\cdot)}$: } It is indeed a subject of ongoing discourse to ascertain whether these automatic evalution metrics such as NPMI or $C_v$ genuinely align with the intricacies of human evaluation criteria~\cite{hoyle2021automated,hoyle2022neural,doogan2021topic}, challenging whether the target of interpretability in Contratopic is ambiguous. However, there are also several works pointing out that these automated coherence metrics are still meaningful. For instance, recent research~\cite{lim2023large} attested to the significance of coherence metrics like NPMI across a spectrum of datasets, encompassing both general corpora such as Wiki and specialized ones like ArXiv and Pubmed. Their findings substantiated that NPMI remains a meaningful automatic coherence metric, showcasing a substantial correlation with human cognition.

The rationale behind our selection of NPMI lies in its established status as a widely employed coherence metric within topic modeling. Moreover, the incorporation of mutual information estimation resonates with our contrastive term's objectives: minimizing mutual information between distinct topics while simultaneously maximizing mutual information within individual topics.

\subsection{Sampling Strategy}
The remaining problem is how to sample $T_k$ from topic $k$, given the topic-word distribution $\beta_k$ calculated by the word embeddings $\rho$ and topic embedding $t_k$. 
However, sampling from a discrete distribution is non-differential.
The Gumbel-softmax trick~\cite{jang2016categorical} provides a simple and efficient way to parameterize a discrete distribution and allow for the propagation of gradients in the sampling process.
In detail, the Gumbel-softmax estimator gives an approximate one-hot sample $\mathbf{y}=\{y_1,y_2,..,y_n\}$ with Eq.\ref{eq:gumbel}.
\begin{equation}\label{eq:gumbel}
y_i=\frac{\exp \left(\left(\log w_i+g_i\right) / \tau_g\right)}{\sum_{j=1}^V \exp \left(\left(\log w_j+g_j\right) / \tau_g\right)},
\end{equation}
where $\tau_g$ is the temperature and  $\log(w_i)$ are logits for a softmax distribution $p(x_i|\boldsymbol{w})=w_i / Z$ and $Z$ is the partition function. 
$y_i$ denotes the probability of random variable $x_i$ being sampled from the discrete distribution in the Gumbel-Softmax estimator.
$g_i=- \log (-\log u_i)$ is called a Gumbel random variable and $u_i \sim \mathrm{Uniform}$    (0,1).
Since the randomness $g$ is independent of $w$, the reparameterization trick can be leveraged to optimize the model with back-propagation.

As we are interested in sampling a subset of words from the topic-word distribution, i.e., drawing samples without replacement, we employ a relaxed subset sampling algorithm proposed by~\cite{xie2019reparameterizable}.
Given the topic-word distribution $\boldsymbol{\beta}$, consisting of $\mathbf{\beta}_k$, and Gumbel variables $\mathbf{g}_k$, a Gumbel-max key is computed for each word as $\mathbf{\hat{r}}_{k}=\log \mathbf{\beta}_k+\mathbf{g}_k$, where $\mathbf{r}_k, \mathbf{\beta}_k, $and $ \mathbf{g}_k \in \mathcal{R}^V$.
A relaxed subset sample of the items can be drawn by applying a relaxed top-v procedure directly on $\hat{r}_k$.
The procedure defines 
\begin{equation}\label{eq:topv1}
\mathbf{r}_{k}^{j+1} := \mathbf{r}^j_k+\log (1-p(\mathbf{r}^j_k=1)),
\end{equation}
where $\mathbf{r}^1_k:=\mathbf{\hat{r}}_k$ and
\begin{equation}\label{eq:topv2}
p(\mathbf{r}^j_k=1)=\frac{\mathrm{exp}(\mathbf{r}^j_k/\tau)}{\sum^V_{m=1}\mathrm{exp}(\mathbf{r}^j_k / \tau)}.
\end{equation}

Finally, a relaxed $v$-hot vector for topic $k$, representing the $v$ words sampled can be computed as $\mathbf{y}_k=\sum^v_{j=1}p(\mathbf{r}_k^j=1)$.

\textbf{Balance between Positive and Negative Samples: }To maintain the simplicity of our method, we abstain from introducing any components aimed at balancing the quantity of positive and negative samples. Nevertheless, our approach does entail the sampling of "v" top words from each topic, which results in 
$k \cdot C^{2}_v$positive word pairs and $v^2 \cdot C^{2}_k$negative word pairs in all the possible word pairs. The balance between these pairs can be modulated by tuning $v$, thereby achieving a proportionate ratio. Of course, other methods such as incorporating a hyper-parameter to balance the weights of negative word pairs can also be considered if necessary.

\subsection{Training Objective}
Our final training objective is 
\begin{equation}
    \mathcal{L}_{tr}=\mathcal{L}_{rec}+\mathcal{L}_{kl}+\lambda\mathcal{L}_{con},
\end{equation}
where $\lambda$ is a hyperparameter.
The training procedure is described in Algorithm \ref{alg:training}.

A possible concern on Contratopic could be whether optimizing evaluation metrics in training objectives is valid or perfectly fair.
Our objective function encompasses not solely the contrastive term, intricately tied to the evaluation metrics, but also integrates the conventional KL-divergence term and reconstruction term. 
The efforts to make the evaluation metrics to be incorporated in the training objectives are also parts of our contribution.
Moreover, optimizing an objective closely aligned with the evaluation metrics is not rare in the domain of machine learning. For instance, the ubiquitous cross-entropy objective function maintains a significant correlation with diverse evaluation metrics, including AUC scores and accuracy measurements. Besides, there are also works in image classification that focus on directly optimizing the AUROC and AUPRC scores during the training stage~\cite{qi2021stochastic,liu2019stochastic}.

\begin{algorithm}[t]
    \caption{The training procedure.}
    \label{alg:training}

\begin{algorithmic}
    \STATE {\bfseries Input:} the input corpus $\mathcal{D}$, topic number $K$, total epoch number $T$, hyperparameter $\lambda$, $v$, temperatures $\tau_g$, $\tau_{\beta}$
    \STATE {\bfseries Output:} $K$ topic-word distributions $\beta_k$, $D$ document-topic distribution $\theta_d$ 
    \STATE Initialize model and variational parameters
    \FOR{$epoch$ from 1 to $T$}
        \FOR{ a random batch of $B$ documents}
            \STATE $\mathcal{L}_{batch} \leftarrow 0$;
            \STATE $\boldsymbol{\beta} \leftarrow \mathrm{softmax}(\mathbf{\rho} \boldsymbol{t}/ \tau_{\beta})$ 
            \FOR{each document $d$ in the batch}
                \STATE $\theta_d \leftarrow q(\theta|\boldsymbol{w_d})$
                \STATE $p(\boldsymbol{w}|\theta,\rho,\boldsymbol{t}) \leftarrow \theta_d ^ \top \boldsymbol{\beta}  $.   
                \STATE $\mathcal{L}_{batch} \leftarrow \mathcal{L}_{batch}+(\mathcal{L}_{\mathrm{rec}}+\mathcal{L}_{\mathrm{kl}})$ by Eq.\ref{eq:elbo};
            \ENDFOR
            %\STATE Estimate $\mathcal{L}_{rec}$ and $\mathcal{L}_{kl}$ by Eq.\ref{eq:elbo}
            \FOR{$k$ from 1 to $K$}
                \STATE $\mathbf{\hat{r}}_{k}\leftarrow\log \mathbf{\beta}_k+\mathbf{g}_k$  
                \FOR{$j$ from 1 to $v$}
                    \STATE Compute $p(\mathbf{r}_k^j=1)$ by Eq.~\ref{eq:topv1} and Eq.~\ref{eq:topv2}
                    %\STATE sample a subset $T_k$ of $v$ words from $\beta_k$
                \ENDFOR
                \STATE $\mathbf{y}_k\leftarrow\sum^v_{j=1}p(\mathbf{r}_k^j=1)$
                \STATE Generate the subset $T_k$ with the $v$-hot vector $\mathbf{y_k}$
            \ENDFOR
            \STATE $\mathcal{L}_{batch} \leftarrow \mathcal{L}_{batch}+\lambda \mathcal{L}_{\mathrm{con}}$ by Eq.\ref{eq:contrastive};
            %\STATE Estimate $\mathcal{L}_{con}$ by Eq.\ref{eq:contrastive}
            \STATE Update model parameters with $\nabla \mathcal{L}_{batch}$
        \ENDFOR
    \ENDFOR

\end{algorithmic}
\end{algorithm}

\subsection{From Mutual Information Estimation Perspective}
The loss function of contrastive learning in our ContraTopic is similar to most mutual information neural estimation methods such as MINE~\cite{belghazi2018mutual}.
However, there are still some differences during the optimization.
MINE uses a lower bound to the mutual information based on the Donsker-Varadhan representation of the KL-divergence and chooses $\mathcal{F}$ to be the family of functions $K_{\phi}: \mathcal{X}\times \mathcal{Z} \xrightarrow{} \mathcal{R}$ parameterized by a deep neural network with parameters $\phi \in \Phi$,
\begin{equation}\label{eq:mine}
I(X;Z) \geq I_{\Phi}(X, Z):=\sup _{\phi \in \Phi} \mathbb{E}_{\mathbb{P}_{X Z}}\left[K_\phi\right]-\log \left(\mathbb{E}_{\mathbb{P}_X \otimes \mathbb{P}_Z}\left[e^{K_\phi}\right]\right),
\end{equation}
where X and Z are random variables and $\mathbb{P}_{XZ}$ is the joint distribution and $\mathrm{P}_X \otimes \mathbb{P}_Z$ is the product of their marginal distribution.
Given the joint and marginal distribution, MINE optimizes the $K_{\phi}$ to maximize the objective, which measures the similarities between samples from X and Z.

In most mutual information neural estimation scenarios, the marginal distribution of variables X and Z is typically available, while the similarity measure network $\mathcal{K(\cdot)}$ is unknown and necessitates parameterization and optimization.
However, in the case of ContraTopic, the similarity measure network $\mathcal{K(\cdot)}$ is pre-defined as the NPMI score between two given words and cannot undergo optimization during training.
In contrast, we parameterize the marginal distribution of X and Z and optimize their distributions to maximize the objective.
In summary, to enhance the coherence and diversity of topics, we employ parameterization and optimization techniques by optimizing the topic-word distribution based on provided similarity measure networks and NPMI scores, thereby maximizing the mutual information within a single topic and minimizing that across different topics.

\begin{table*}[h]
\centering

\caption{Summary of the statistics for three datasets}
\label{tab:stat}
% \begin{tabular}{l|ccc}
% \toprule
%                         & 20NG & Yahoo & NYTimes    \\
% \midrule
% Vocabulary Size       & 5,770       & 7,394      & 34,330     \\
% Training Samples         & 10,827       & 89,808       & 179,814     \\
% Test Samples          & 7,183      & 59,873      & 119,876    \\
% Average Length & 59.8      &     45.9  &345.7      \\
% Number of Tokens      & 1,076,941     &6,872,000    &103,608,732  \\
% \bottomrule
% \end{tabular}

\begin{tabular}{l|ccccc}
\toprule
& Vocabulary Size& Training Samples &Test Samples&Average Length&Number of Tokens\\
\midrule
20NG       & 5,770       & 10,827      & 7,183&59.8&1,076,941     \\
Yahoo         & 7,394       & 89,808       & 59,873& 45.9&6,872,000     \\
NYTimes         & 34,330      & 179,814      & 119,876&345.7&103,608,732    \\
\bottomrule
\end{tabular}

\vspace{-0.1in}
\label{tab:data_statistics}
\end{table*}

\subsection{Differences between ContraTopic and Other methods}
A pivotal distinction between our method and other topic models that incorporate contrastive terms stems from the foundational insights that underscore each approach. As delineated in related works, other methods, exemplified by CLNTM~\cite{nguyen2021contrastive}, employ a "document-wise" contrastive term, with their guiding principle centered around nurturing similar document-topic distribution among similar documents, which will implicitly benefit the topic-word distribution and thereby contribute to topics with higher quality. In stark contrast, our approach adopts a "topic-wise" contrastive term, which, in a more direct and unequivocal manner, fosters coherence and diversity of the topic-word distribution, leading to an enhancement in the quality of topics. Empirical experiments corroborate this distinction, revealing that our topic-wise contrastive term eclipses the quality achieved by CLNTM, thus underscoring the tangible merits of optimizing the topic-word distribution to improve topic quality.

Compared with CLNTM, our definitions of samples, noted as $i$ or $a$ in eq(2), are different. In Contratopic, we treat predominant words from topics as samples while CLNTM treats documents as samples. Such differences mainly result from the different fundamental insights discussed above.

Besides, the definitions of similarity functions, noted as $\mathcal{K}(\cdot)$ in eq(2), are also different. In Contratopic, we use the precomputed NPMI scores to measure the similarity between words while CLNTM leverages dot product among document representations to measure the similarity between documents.

Last but not least, the sampling strategies, noted as $P(i) $ and $ I $ in eq(2), for generating positive and negative samples are still different. CLNTM indeed leverages a word-based sampling strategy to draw positive and negative documents via modifying the weights of significant tokens based on the tf-idf scores. However, our positive and negative words are defined by the topics they sampled from, which is completely different.

%互信息 传统对比学习的区别 EVALUATION NTMR
\section{Experiments}
\subsection{Datasets}
Our experiments are conducted on three widely-used datasets, including 20 newsgroups (20NG), UIUC Yahoo Answers (Yahoo), and New York Times (NYTimes).
20NG~\cite{lang1995newsweeder} is a dataset that contains around 20,000 newsgroup documents and is commonly used in the topic modeling field. 
Yahoo~\cite{chang2008importance} is a larger dataset that contains nearly 150,000 documents related to lots of questions and corresponding answers.
NYTimes~\footnote{https://archive.ics.uci.edu/ml/datasets/Bag+of+Words} is a collection of about 300,000 news articles with a huge vocabulary.
We preprocess documents in each dataset by tokenizing, filtering out stop words, words with document frequency above 70\%, and words appearing in less than around 100 documents (depending on the dataset). Then we remove the documents shorter than two words. 
It should be noted that 20NG and Yahoo are associated with document labels.
We follow the original division of training and test sets on 20NG.
For Yahoo and NYTimes, the ratio of training to test samples is 6:4.
The statistic details of our used datasets are shown in Table~\ref{tab:data_statistics}.

% The detailed instructions of preprocessing and statistics of each dataset are shown in Appendix. 

% Below, we present a list of preprocessing steps for easy reference.
% \begin{compactenum}
%     \item[] Document processing
%         \begin{compactenum}
%             \item[] a. We perform basic processing such as stopword removal to all the documents;
%             \item[] b. Following processing, we remove documents with fewer than two tokens.;
%             \item[] c. Given that we employ the bag-of-words representation for each document, we do not truncate excessively lengthy documents.
%         \end{compactenum}    
%     \item[] Vocabulary creation 
%         \begin{compactenum}
%             \item[] a. We tokenize using \textit{CountVectorizer} in \textit{sklearn}.;
%             \item[] b. We lowercase terms.;
%             \item[] c. We do not lemmatize.;
%         \end{compactenum}
%     \item[] Vocabulary filtering 
%         \begin{compactenum}
%             \item[] a. The vocabulary is created from the training and test data. The test texts used in coherence calculations are processed identically and use the same vocabulary.;
%             \item[] b. We retain only tokens that are matched by the regular expression
%          ˆ[\w-]*[a-zA-Z][\w-]*\$.;
%             \item[] c. We remove tokens that appear in more than 70\% of documents.;
%             \item[] d. We remove tokens that appear in fewer than around 100 documents.;
%         \end{compactenum}
% \end{compactenum}

\begin{figure*}[t]
\centering
  \includegraphics[width=6 in]{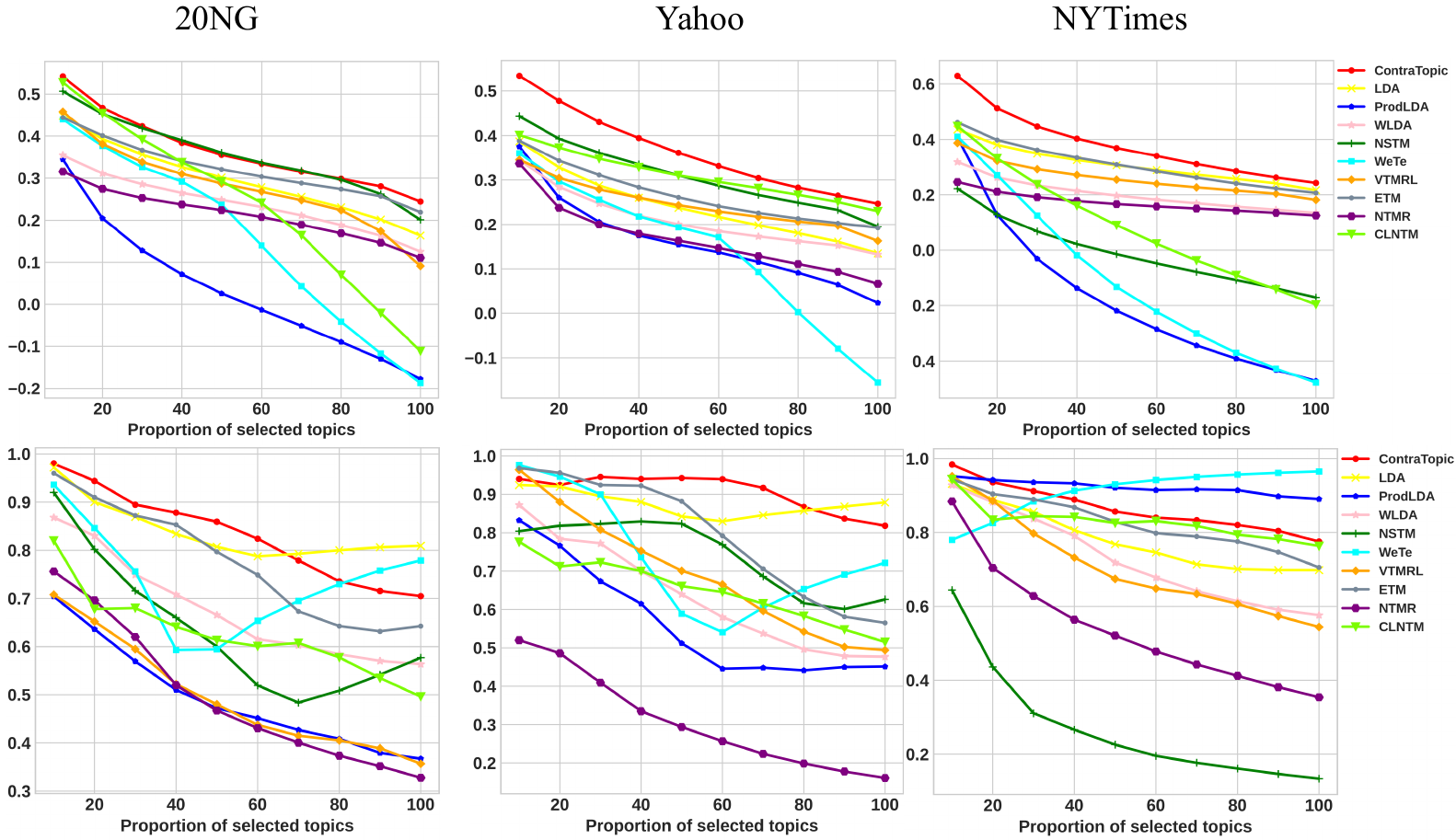}
  \caption{The results of topic interpretability evaluation. The first row shows the topic coherence of all the datasets in the test set, respectively. The second row shows the corresponding topic diversity scores. In each subfigure, the horizontal axis indicates the proportion of selected topics according to their NPMIs. }
  \label{fig:topic}
\end{figure*}

\subsection{Evaluation Metrics}

For topic interpretability, we choose two kinds of common metrics: topic coherence, and topic diversity.
\textbf{Topic coherence} measures the average NPMI over the topic $K_{TC}$ words of the selected topics.
\textbf{Topic diversity} measures the percentage of unique words in the top $K_{TD}$ words of selected topics.
To comprehensively evaluate the topics, we follow the experimental settings in NSTM~\cite{zhao2020neural} and report the average NPMI scores in different percentages of topics.
The proportions of the selected topics vary from 10\% to 100\%.
Following previous works~\cite{zhao2020neural,wang2022representing}, $K_{TC}$ is set to 10 and $K_{TD}$ is set to 25.

% Following ~\cite{hoyle2021automated}, we also conduct human evaluation, a word intrusion task, to assess the topic interpretability in a behavioral way.
% The core idea of the word intrusion task is that when the top words in a topic identify a coherent latent category, it is easier to identify words that do not belong to that category.
% Each topic is represented as its top words plus one "intruder" word.
% To control the workload of human annotators, we only show the top 5 words of each topic and evenly sample 10\% topics with different NPMI.
% We invite 12 graduate students as annotators.
% All participants are told that they are performing human evaluations for an automatic method, and none of their private information is collected during the experiments.
% An example of the questionnaire used in the human evaluation is shown in the Appendix.
% We generate the "intruder" word from top words in other topics so that both the coherence and diversity of topics can be evaluated.
% The \textbf{word intrusion scores (WIS)}, varying from 0 to 1, measures how well human annotators detect the "intruder" word.

For the evaluation of document-topic distributions, we perform document clustering tasks.
We report the purity and Normalized Mutual Information (NMI) on 20NG and Yahoo, where the document labels are available.
%计算公式
Given the document-topic distribution as the document representation, we apply the KMeans algorithm on test data and report the scores of the KMeans clusters (denoted by \textbf{km-Purity} and \textbf{km-NMI}), following~\cite{wang2022representing}.
The number of clusters in KMeans varies in the range of {20, 40, 60, 80, 100}.
For all the metrics, higher values indicate better performance.

\begin{figure*}[t]
\centering
  \includegraphics[width=6.5 in]{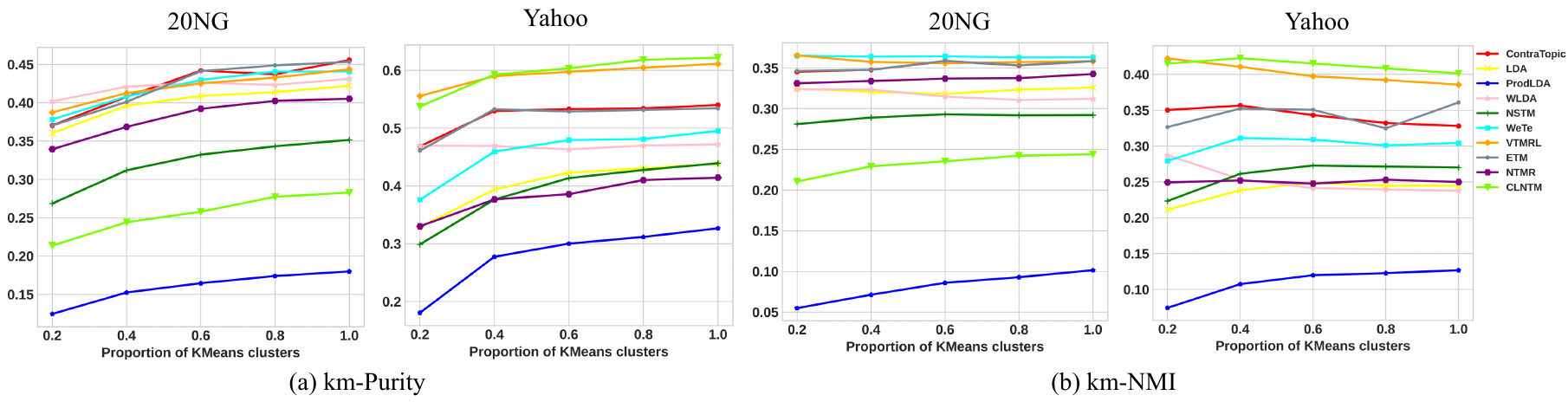}
  \caption{The results of document representation evaluation. (a) The two subfigures show the km-Purity scores on 20NG (left) and Yahoo(right). (b) The two subfigures show the km-NMI scores on 20NG (left) and Yahoo (right). }%The horizontal axis indicates the percentage of cluster numbers relative to the number of topics. }
  \label{fig:doc}
\end{figure*}
\begin{table*}[t]
\centering

% \scalebox{0.80}{\begin{tabular}{l|ccc}
% \toprule
% Methods&topic coherence&topic diversity&km-Purity\\
%  \midrule
% ContraTopic   & 0.24 - \textbf{0.51}      &0.73 - \textbf{0.98}       &\textbf{0.41} - \textbf{0.48}\\ \hline
% ContraTopic-P & \textbf{0.26} - 0.43          & 0.68 - \textbf{0.98}          & \textbf{0.41} - 0.47              \\
% ContraTopic-N & 0.12 - 0.41          & 0.60 - 0.96          & 0.34 - 0.38                \\
% ContraTopic-I & 0.23 - 0.44          & 0.70 - 0.95        & 0.40 - 0.46                \\
% ContraTopic-S & 0.22 - 0.45          & \textbf{0.74} - 0.96         & 0.38 - 0.45               \\

% % SeededNTM-topic & 0.327          & 0.553          & \textbf{0.561} \\ 
% \bottomrule
% \end{tabular}}
\caption{Results of different variants on 20NG. }
\begin{tabular}{l|ccc|ccc|ccc}

\toprule
Metrics & \multicolumn{3}{c|}{Topic Coherence}                                     & \multicolumn{3}{c|}{Topic Diversity}                                    & \multicolumn{3}{c}{km-Purity}                                       \\
  Percentage                   & 10\%          & 50\%           & 90\%            & 10\%          & 50\%           & 90\%           & 20\%          & 60\%           & 100\%            \\ 
\midrule
ContraTopic              & \textbf{0.54$\pm$0.2}           & \textbf{0.36$\pm$0.1}          &\textbf{ 0.28$\pm$0.3 }          & \textbf{0.98$\pm$0.1}           &\textbf{ 0.86$\pm$0.2}           & \textbf{0.72$\pm$0.2}          & \textbf{0.37$\pm$0.0}           & 0.44$\pm$0.1           & \textbf{0.46$\pm$0.0}              \\ \hline
ContraTopic-P            & 0.44$\pm$0.2           & 0.33$\pm$0.1          & 0.27$\pm$0.1           & 0.98$\pm$0.1           & 0.83$\pm$0.2           & 0.69$\pm$0.1          & 0.36$\pm$0.0           & \textbf{0.45$\pm$0.0}           & 0.44$\pm$0.0   \\
ContraTopic-N            & 0.42$\pm$0.2         & 0.27$\pm$0.4          & 0.19$\pm$0.5           & 0.95$\pm$0.1           & 0.69$\pm$0.3           & 0.61$\pm$0.2          & 0.34$\pm$0.1           & 0.37$\pm$0.1           & 0.38$\pm$0.1                 \\
ContraTopic-I            & 0.45$\pm$0.1         & 0.33$\pm$0.2          & 0.26$\pm$0.2           & 0.95$\pm$0.1           & 0.84$\pm$0.3           & 0.70$\pm$0.2          & 0.35$\pm$0.0           & \textbf{0.45$\pm$0.0}           & 0.44$\pm$0.0                \\
ContraTopic-S            & 0.50$\pm$0.2         & 0.34$\pm$0.1          & 0.26$\pm$0.1          & 0.96$\pm$0.1           & 0.85$\pm$0.2        & \textbf{0.72$\pm$0.1}          & 0.36$\pm$0.1           & 0.44$\pm$0.0           & 0.45$\pm$0.0                 \\ 
\bottomrule
\end{tabular}

\label{tab:ablation}
\end{table*}
\subsection{Experimental Settings}
We compare the performance of our proposed model with the following baselines: \textbf{1)} conventional topic models, LDA~\cite{blei2003latent}, the most famous topic model; \textbf{2)} VAE-based NTMs, such as ProdLDA~\cite{srivastava2017autoencoding}, which replaces the mixture model in LDA with a product of experts, and WLDA~\cite{nan2019topic}, which replaces the KL-divergence with the Maximum Mean Discrepancy, and ETM~\cite{dieng2020topic}. which incorporates word embeddings into NTMs; \textbf{3)} OT-based NTMs, NSTM~\cite{zhao2020neural}, which learns the topics proportions by directly minimizing the OT distance to the document-word distribution, and WeTe~\cite{wang2022representing}, which views each document as a set of word embeddings and projects topics into the same embedding space; \textbf{4)} NTMs that incorporate topic interpretability into objectives, such as NTM-R~\cite{ding2018coherence}, which designs a topic coherence objective into the training process and VTMRL~\cite{gui2019neural}, which incorporates topic coherence measures as reward signals. 
\textbf{5)} NTMs that also leverage contrastive learning but in a document-wise manner, such as CLNTM~\cite{nguyen2021contrastive} which encourage similar document-topic distribution among related documents.
%\textcolor{red}{Note that we do not compare with baselines related to contrastive learning since they focus on improving the document-topic representations and can be combined with our method.}
%deals with different problem settings and can also be used together with our regularizer.}
For all the above baselines, we either use their official code or reproduce their models with the best reported settings.

\subsection{Settings for Our Proposed Model}
We implement ContraTopic on Pytorch.
We use ETM as our backbone model and share the same hyper-parameters.
For the encoder, we use a three-layer perceptron of 800 hidden units and SeLU as the activation function, followed by a dropout layer (rate = 0.5) and a batch norm layer.
Similar to several baselines, we use the GloVe word embeddings pre-trained on Wikipedia\footnote{ https://nlp.stanford.edu/projects/glove/}.
We freeze the word embeddings during the training time for stability.
The number of topics is set to 100 in all the topics.
We use the Adam optimizer with a learning rate of 0.0005 and batch size of 1000 for 100 epochs.
For our regularizer, we implement the function $\mathcal{K}(\cdot)$ with the pre-computed NPMI scores on the training set.
Therefore, we evaluate the topic coherence on the \textit{unseen test data} to make fair comparisons.
Since our method only adds a regularizer to the backbone model, we keep the shared hyper-parameters unchanged and perform the grid search for other hyper-parameters such as $\lambda, v, \tau_g,$ and $\tau_g$ on a validation set split from the training corpus. 
The $\lambda$ is set to 40, 40, and 300 for 20NG, Yahoo, and NYTimes respectively.
Other hyper-parameters such as $v, \tau_\beta$, and $\tau_g$ are set to the same among three datasets where $v$ is set to $10$, $\tau_\beta$ is set to $0.1$ and $\tau_g$ is set to be $0.5$.
The code and our used 20NG dataset can be found in supplementary materials.

\subsection{Computational Analysis}
\textbf{Theoretically}, the increased time and space requirements in the proposed approach can be attributed to two factors: the sampling process and the pair-wise computation of NPMI in $\mathcal{K(\cdot)}$. The sampling process primarily contributes to additional time costs, with a complexity of O($M$) where $M$ represents the number of samples. On the other hand, the pair-wise computation of NPMI leads to increased spatial costs, specifically O($V^2$), as it involves storing the precomputed NPMI matrix. Additionally, the precomputation of NPMI can be performed during preprocessing, requiring a time equivalent to approximately 30 training epochs.

\textbf{Empirically}, our experiments, conducted on a machine equipped with 2 Nvidia RTX8000 GPUs and Intel Xeon Gold 6230 CPUs, show that ContraTopic, \textbf{even with the complete NPMI matrix in GPU memory}, requires 14593MiB GPU memory (8673MiB GPU memory if we only maintain the matrix in CPU memory) and 65.68 sec/epoch for NYTimes, a relatively modest resource consumption. 
We believe the extra computational expense is justified by the improvements.

\begin{figure*}[h]
\centering
  \includegraphics[width=6.5 in]{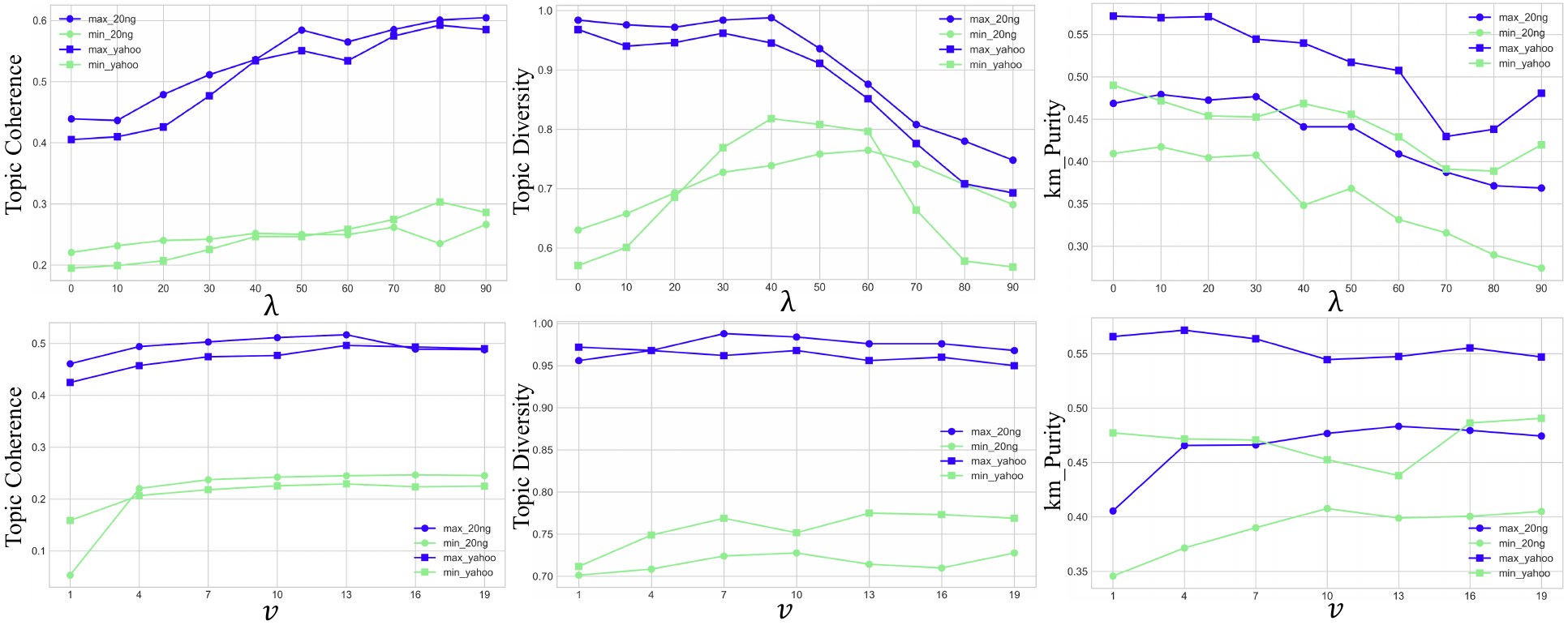}
  \caption{The sensitivity analysis results of $\lambda$ and $v$ on 20NG and Yahoo. }
  \label{fig:sensitivity_1}
\end{figure*}
\begin{figure*}[h]
\centering
  \includegraphics[width=6.5 in]{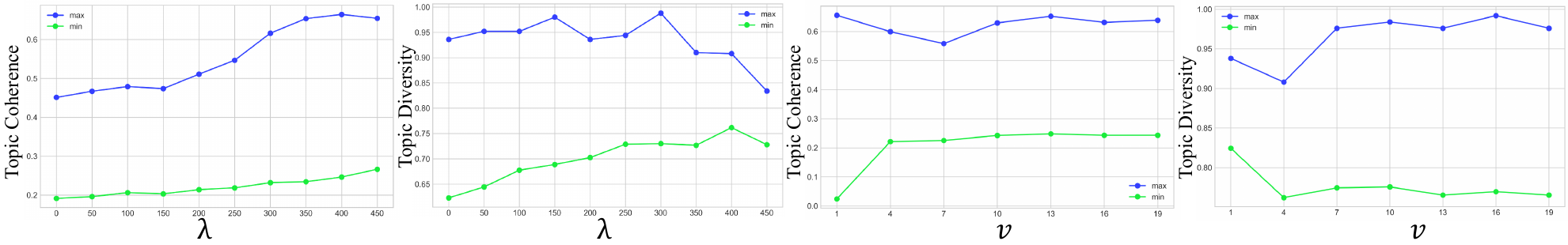}
  \caption{The sensitivity analysis results of $\lambda$ and $v$ on NYTimes. }
  \label{fig:sensitivity_2}
\end{figure*}

\subsection{Results}
The results of topic coherence and topic diversity on three datasets are shown in Figure~\ref{fig:topic}.
We run each model in comparison three times by modifying only the random seeds and reporting the mean values. 
The error bars are omitted in the figure for brevity.
From the results, we can observe that ContraTopic outperforms almost every baseline in terms of topic coherence.
Though NSTM has competitive results on 20NG, its topic diversity still has a certain gap with ContraTopic.
\textbf{Some baselines have shown an obvious decline in topic coherence but an increment in topic diversity when the evaluated topics increase.}
A possible reason is that the topics in the tail are of very low quality containing some infrequent words that are quite different from the words in topics with high NPMI.
Despite achieving a high level of topic diversity in NYTimes, both WeTe and ProdLDA exhibit a significant decline in their topic coherence, which falls considerably short of those achieved by ContraTopic.
% The results of word intrusion scores are provided in Table~\ref{tab:intrusion}.
% ContraTopic and NSTM have competitive results, while ETM and WeTe also have better results compared with other baselines, indicating topics generated by ContraTopic are easier for humans to understand.
% Some topics on three datasets are provided in Appendix.
% We also substitute ContraTopic’s backbone model from ETM to WLDA and WeTe to
% validate ContraTopic’s effectiveness across various architectures, which can be found in Appendix.
\begin{figure*}[h]
\centering
  \includegraphics[width=6 in]{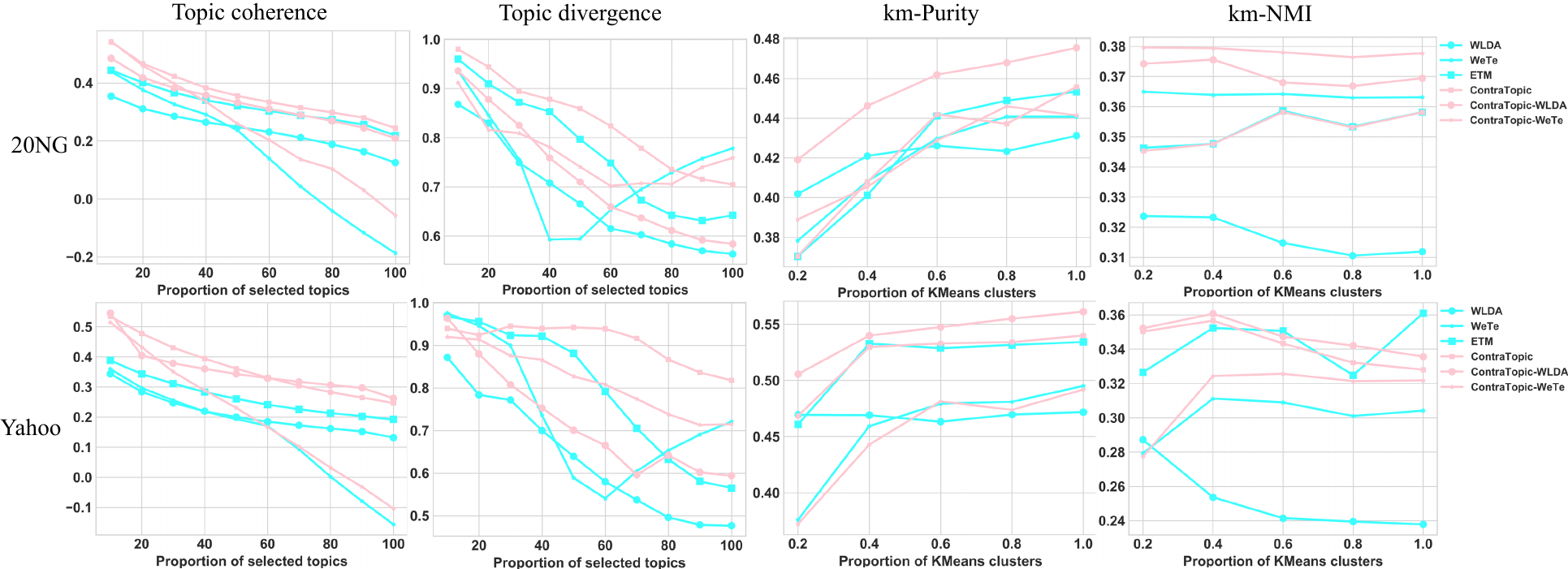}
  \caption{
The results of backbone model substitution for 20NG in the first row and Yahoo in the second row. The blue lines correspond to various backbone models, while the pink lines represent backbone models improved by our regularizer. Matching markers indicate the usage of the same backbone models.}
  \label{fig:backbone}
\end{figure*}

Figure~\ref{fig:doc} presents the results of km-Purity and km-NMI. Despite not incorporating any specific techniques for document representation, ContraTopic manages to achieve competitive outcomes on 20NG.
Although ETM and VTMRL outperform ContraTopic on Yahoo, they fall short in terms of topic coherence and diversity. 
\textbf{Besides, we have to posit that the results obtained from km-Purity and km-NMI, despite their application in numerous studies, merely encapsulate a fraction of the multifaceted nature inherent in document representation quality.} These metrics essentially measure the alignment between unsupervised document-topic distributions and human-annotated labels. Nonetheless, the alignment between the generated topics in unsupervised topic models and the annotated labels of the dataset remains uncertain. Consider, for instance, a toy dataset comprising nine news articles, each centered around politics, sports, and economics within the United States, Japan, and China. In the course of manual annotation, we designated three articles for each category: "politics," "sports," and "economics." When subjecting these articles to topic modeling, there is no assurance that the acquired topics will precisely correspond to politics, sports, and economics. It is plausible that due to the involvement of common public figures in news from the same nation, the model might discern "United States," "Japan," and "China" as three distinct topics. In such a scenario, even if we obtain high-quality topics, the model may still achieve low scores in km-Purity and km-NMI. As a consequence, we posit that a lower score in km-Purity and km-NMI does not mean worse document representation but the inconsistency between generated topics and annotated dataset labels.

In summary, ContraTopic excels in extracting high-quality, interpretable topics while also preserving the capability to learn useful document representations for clustering.

\subsection{Ablation Study}

The results of our ablation study are shown in Table~\ref{tab:ablation}. The percentage values for topic coherence and topic diversity indicate the proportion of selected topics, whereas the percentage values for km-Purity represent the proportion of clusters.

\textbf{Contrastive objective.} 
The crucial design decision in ContraTopic centers around the contrastive learning objective denoted as $\mathcal{L}_{con}$, which can be broken down into two components: the calculation involving positive sample pairs and negative sample pairs. To assess the impact of each component, we introduce two variants: 1) ContraTopic-P, which solely focuses on positive sample pairs, and 2) ContraTopic-N, which solely focuses on negative sample pairs. The performance of topic coherence and diversity exhibits a decline of approximately $5\%$ for ContraTopic-P and $12\%$ for ContraTopic-N. This finding underscores the significance of considering multiple facets of interpretability. Comparatively, the km-Purity remains competitive for ContraTopic-S, while ContraTopic-N experiences a significant deterioration. These results indicate that exclusively considering negative sample pairs is detrimental to the performance of NTMs.

\begin{table*}[h]
\centering

\caption{Results of the word intrusion scores on 20NG}
\label{tab:intrusion}

\begin{tabular}{c|ccccccccc|c}
\toprule
&LDA&ProdLDA&WLDA&ETM&NSTM&WeTe&NTMR&VTMRL&CLNTM&ContraTopic          \\
\midrule
\textbf{WIS} &0.34 & 0.37&0.34&0.58&0.68&0.67&0.29&0.46&0.64& \textbf{0.80}    \\
\bottomrule
\end{tabular}

\label{tab:intrusion}
\vspace{-0.15in}
\end{table*}

\textbf{Similarity measure network.} Another core component of ContraTopic is the similarity measure network which is implemented as pre-defined NPMI scores. 
We design a variant following NTM-R~\cite{ding2018coherence} to implement $\mathcal{K}(\cdot)$ with the inner product between word embeddings, noted as ContraTopic-I.
We find that ContraTopic substantially outperforms this variant, suggesting the advantage of using NPMI for $\mathcal{K}(\cdot)$, which aligns closer with the current evaluation of topic interpretability.

\textbf{Sampling strategy.} The final key design of ContraTopic is that we introduce a sampling process in the estimation of mutual information between topics.
A possible variant is to leverage the weight sum operation of topic-word distribution as an expectation for the mutual information estimation, noted as ContraTopic-S.
We find that ContraTopic-S has the least performance decline.
The result suggests that the sampling strategy helps the model to improve topic interpretability.

% We analyze the effects of different modules in ContraTopic by comparing among the following variants: 1) ContraTopic-P: only considering positive samples; 2) ContraTopic-N: only considering negative samples; 3) ContraTopic-I: implementing the function $\mathcal{K}(\cdot)$ with the inner dot between word embeddings; 4) ContraTopic-S: replacing the $v$-hot vector in sampling strategy with the topic-word distribution vector to perform the weight sum operation, which is similar with~\cite{ding2018coherence}.

% Performances on 20NG are provided in Table~\ref{tab:ablation}.
% Different from the results in Figure~\ref{fig:topic}, we only provide the minimal and maximal values in the curves.
% Besides, the results of km-NMI are not shown in the table since they are similar to km-Purity.
% Except for the decline in topic diversity, ContraTopic-P has almost competitive results with ContraTopic, which matches our intuition that encouraging negative samples improves topic diversity.
% However, only considering negative samples leads to a large decline in topic diversity and km-Purity, which proves our intuition to take into multiple aspects of topic interpretability.
% The results of ContraTopic-I and ContraTopic-S show that our proposed similarity function and sampling strategy have positive effects compared with other variants.

\subsection{Sensitivity Analysis }
The sensitive analysis is conducted on two significant hyper
parameters, $\lambda$, controlling the power of our regularizer and $v$, the number of words sampled for topic-word distribution.
Similar to Table~\ref{tab:ablation}, we only report the highest and lowest results (under the maximum and minimum percentages) of topic coherence, topic diversity, and km-Purity.
The results for other percentages exhibit a consistent trend, thus we exclude them for the sake of conciseness.
The results are provided in Figure~\ref{fig:sensitivity_1} and Figure~\ref{fig:sensitivity_2}.
We plot the results of 20NG and Yahoo in the same subfigure due to space limitations.
Besides, the scale of $\lambda$ in the NYTimes is also much larger than the other two datasets.
The trends on all three datasets are similar.

When we vary $\lambda$ from 0 to 90, the topic coherence increases gradually, especially for most coherent topics.
As for the diversity, the maximal and minimal scores increase at first and decline when the $\lambda$ gets too large.
The km-Purity has a similar tendency with topic diversity.
Interestingly, when $\lambda$ gets larger, the topic diversity, though declining, seems to have a smaller variance among topics.
As we vary the parameter $v$ within the range of 1 to 19, we observe a rapid initial increase in both topic coherence and km-Purity, followed by a sustained high level.
We found the choice of $\lambda$ is more sensitive to different datasets while $v$ seems to be less sensitive.
Empirically, choosing an appropriate $\lambda$ is important for different datasets.
We believe our sensitivity analysis can help readers apply our method to their own dataset.

A discernible trade-off emerges between topic coherence and topic diversity when $\lambda$ gets large values, a phenomenon acknowledged by other studies as well~\cite{wu2020short}. This observation is potentially attributable to the emergence of multiple repetitive topics with exceedingly high coherence, leading to elevated topic coherence scores but terrible topic diversity scores. Alternatively, the phenomenon may also arise due to the risk of an excessively large lambda overshadowing the original loss function in favor of the contrastive term, thus compromising the integrity of topic modeling.

\subsection{Backbone Model Substitution}
To validate ContraTopic's effectiveness across different architectures, we substitute ContraTopic's backbone model from ETM to WLDA and WeTe.
The results are shown in Figure ~\ref{fig:backbone}.
Our regularizer consistently improves topic coherence and diversity across different backbone models, as indicated by the results. Furthermore, backbone models with our regularizer demonstrate competitive performance in terms of km-Purity and km-NMI. Notably, WLDA benefits significantly from our regularizer, exhibiting substantial improvements in both km-Purity and km-NMI. This observation guides our future exploration towards enhancing both topic quality and document representation quality simultaneously.

\subsection{Human Evaluation}
One primary use of topic models is in computer-assisted content analysis, where topics help humans understand large amounts of text, organize information, retrieve relevant documents, personalize recommendations, detect trends, facilitate knowledge sharing, and gain valuable insights from complex datasets.
As a result, human evaluation is crucial in deriving new topic models because it helps ensure that the models align with human understanding and expectations. Human evaluators can provide subjective judgments, validate the quality of topics generated that automatic evaluation alone may not capture, leading to more accurate and effective topic models.

\subsubsection{Human evaluation Design}
Based on the research conducted by Hoyle et al~\cite{hoyle2021automated}, we conducted a word intrusion task as our human evaluation for topic quality. The word intrusion task involves identifying words that do not belong when the top words in a topic represent a coherent latent category. We invited 20 participants to perform the human evaluation, ensuring their privacy and not collecting any personal information. To evaluate both the coherence and diversity of topics, we generated "intruder" words from the top words in other topics. The word intrusion scores (WIS), ranging from 0 to 1, quantitatively measure the ability of annotators to detect the "intruder" word.

\subsubsection{Questionnaire Generation}
The questionnaire has been completed by all participants.
For each evaluated method, thirty topics were selected from the total of 100 topics in 20NG for evaluating each model, including Contratopic.
Consequently, our questionnaire comprises 300 questions (10 methods * 30 topics per method). Each question requires selecting an intruder word from a set of five top words within a given topic.

\textbf{Topic Selection: }
Due to the impracticality and time constraints of evaluating all topics across each method, we employ a selective approach in human evaluation, focusing on a subset of topics. To ensure fairness in our comparisons, we randomly sample 3 topics from each decile of topics sorted by topic coherence, ensuring even representation across all topics. To avoid bias caused by repeated or similar topics in different models, we iterate the topic selection process until no duplicates are present in a single questionnaire.

\begin{figure}[h]
  \centering
  \includegraphics[width=3 in]{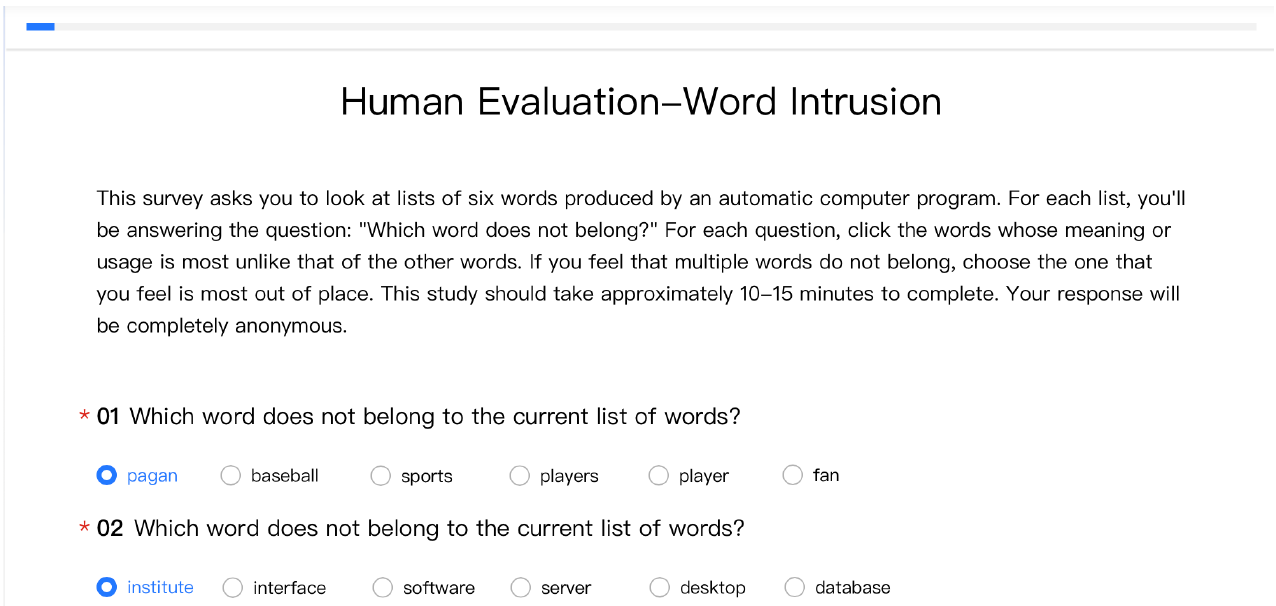}
  \caption{An Example of the questionnaire used in our human evaluation.}
  \label{fig:human}

\end{figure}

\textbf{Intruder Generation: }The intruder word is chosen from topics that were not included in the question set.
To manage the workload of participants, we only chose the five most probable words from each selected topic. Along with these words, we randomly select an intruder word from a pool of words with low probabilities in the current topic (to minimize the chance of it belonging to the same semantic group) but a high probability in other unselected topics (to ensure it is not outright rejected due solely to rarity). All six words are then shuffled and presented to the participant.
An example of our questionnaires is shown in Figure~\ref{fig:human}.

% \textbf{User Instruction: }

\subsubsection{Participants}
We recruit 20 participants including 12 males and 8 females with a computer science background, familiar with basic news concepts and topic models.
The majority of the participants recruited have at least a graduate degree or are undergraduates.
The average age is 26.3.

We do understand that a larger sample size of participants from diverse demographic groups would have increased the statistical power and generalizability of the human evaluation. 
However, our sample size was determined based on available resources and time constraints.
Furthermore, we note that our sample size is still comparable to previous works, as cited in ~\cite{harandizadeh2021keyword} and ~\cite{meng2020discriminative}, which have conducted similar human evaluation experiments that meet scientific standards.
Besides, ~\cite{hoyle2021automated} also shows that a minimum of fifteen crowd-workers per topic for human evaluation is enough to ensure sufficient statistical power.
We pay participants 18 USD for finishing the questionnaire, equivalent to nearly 14 USD/hour.
\subsubsection{Results}
Table~\ref{tab:intrusion} displays the results of our human evaluation. The findings exhibit similarity to the topic coherence results, showcasing ContraTopic's superior performance over all baseline models.
While the selection of intruder words is important, we have observed that participants face greater challenges in correctly identifying intruders within topics with lower coherence. This observation highlights the alignment between automatic and human evaluations of topic models, further strengthening their reliability and consistency.

\begin{table}[t]
\centering

\caption{Generated topics on 20NG.}
\resizebox{\linewidth}{!}{

\begin{tabular}{c|c|l}

\toprule
Models&NPMI&Topic Word Examples\\

 \midrule
\multirow{5}{*}{LDA}    &0.41  &israel jews israeli war jewish arab state land\\
                         &0.37  & space nasa data earth launch orbit moon shuttle\\
                         &0.37  & people \textbf{gun} \textbf{government} armenian armenians turkish \textbf{police guns}\\
                          &0.36 & game team year games season hockey league players\\
                           &0.35 & key encryption chip keys clipper public security government\\
                           \midrule
\multirow{5}{*}{ETM}   &0.67   & armenian armenians turkish people turkey greek turks ottoman\\
                         &0.52  & patients health medical disease cancer drug study drugs  \\
                        &0.43   & space nasa launch orbit earth satellite lunar shuttle\\
                         &0.42  & drive scsi disk hard controller drives bus floppy\\   
                          &0.42 & government people jobs russian work program working money \\
                          \midrule
\multirow{5}{*}{WeTe}  &0.63   &cadre geb pitt surrender gordon skepticism chastity intellect\\
                        &0.46   & disease pain patients drugs medical cancer symptoms blood\\
                        &0.46   & god jesus bible church christian faith christ christians\\
                        &0.41    &armenian turkish armenians people jews war government turkey\\
                         &0.40 & fbi koresh fire news waco police batf compound\\
                         \midrule
\multirow{5}{*}{CLNTM}  &0.68   &\textbf{chastity cadre geb surrender shameful dsl intellect gordon}\\
                        &0.68   &\textbf{shameful geb cadre intellect chastity surrender skepticism gordon}\\
                        &0.68   & \textbf{chastity cadre geb shameful dsl intellect skepticism surrender}\\
                        &0.53  &glenn rice readme solaris oz gcc compatibility ftp\\
                        &0.46    &detroit rangers rockies louis chicago toronto pittsburgh boston\\
 \midrule
\multirow{5}{*}{ContraTopic}  &0.69& armenian armenians turkish armenia genocide azerbaijan turks turkey\\
                        &0.52& health patients medical disease cancer drug diseases study \\
                        &0.50& god jesus church christ christian bible faith holy \\
                        &0.45& server motif application widget export uk applications client \\
                        &0.43& image graphics images jpeg color gif format picture \\
                    \bottomrule 
\end{tabular}}
  \vspace{-0.15in}

\label{tab:20ng}

\end{table}

\subsection{Case Study}
We present the generated topics for 20NG, Yahoo, and NYTimes datasets. 
Topics with the highest NPMI values are shown in Table~\ref{tab:20ng}, ~\ref{tab:yahoo}, and \ref{tab:nytimes}.
We show the top topics generated by our method and some representative baselines such as LDA, ETM, WeTe, and CLNTM.
Besides, with the popularity of large language models, there are works~\cite{stammbach2023re} utilizing large language models to evaluate the quality of generated topics, so we also use large language models to generate a description of our topic to better express the semantic information of each topic based on their topic-word distribution and their most related documents. We only present the top words of the top 5 topics for each method and descriptions for the top 5 topics in ContraTopic due to space limitations.

We can observe that some baselines generate topics that mix with other topics, resulting in a decrease in NPMI, such as the third topic of LDA in 20NG. Besides, some baselines also generate topics that overlap with other topics which leads to a decline in topic diversity, such as the topics generated by ETM on Yahoo. Although the top topics displayed look good, the two types of issues mentioned above often occur for topics with slightly lower NPMI rankings in WeTe. For baselines like CLNTM with high topic consistency and poor topic diversity, there are obvious repetitions in their top topics.

% \begin{table}[h]
% \centering
% \caption{Generated topics on 20NG.}
% \scalebox{0.8}{\begin{tabular}{c|l}
% \toprule
% NPMI&Words\\
%  \midrule
% 0.69& armenian armenians turkish armenia genocide azerbaijan turks turkey soviet\\
% 0.52& health patients medical disease cancer drug diseases study treatment drugs\\
% 0.50& god jesus church christ christian bible faith holy christians heaven\\
% 0.45& server motif application widget export uk applications client user unix\\
% 0.43& image graphics images jpeg color gif format picture software formats\\
% 0.40& space nasa launch orbit earth satellite lunar spacecraft shuttle moon\\
% 0.40& drive scsi disk hard drives controller dos bus floppy ide\\
% 0.38& law government state rights laws court amendment federal legal
% constitution\\ 
% 0.37&  windows screen color video monitor mouse display vga font fonts\\
% 0.37& god religion true truth christian belief religious beliefs exist christianity\\
% \hline 
% \end{tabular}}

% \label{tab:20ng}
% \end{table}

\begin{table}[t]
\centering
\caption{Generated topics on Yahoo.}
\resizebox{\linewidth}{!}{
\begin{tabular}{c|c|l}

\toprule
Models&NPMI&Topic Word Examples\\

 \midrule
\multirow{5}{*}{LDA}    &0.48  &oil cheese sauce pepper garlic juice fresh green\\
                         &0.47  & cup add salt minutes sugar butter mix cream\\
                         &0.45  & weight body fat lose eat healthy diet exercise\\
                          &0.40 & product stores forever asp shoes shirt outfit category\\
                           &0.38 & food eat chicken meat rice eating cook corn\\
                           \midrule
\multirow{5}{*}{ETM}   &0.47   & \textbf{phone} number send email mail \textbf{cell} plan service \\
                         &0.47  & \textbf{food} \textbf{eat} drink milk \textbf{foods} eating meat fruit \\
                        &0.46   & god believe religion church christians christian jesus faith\\
                         &0.44  & ipod music ps song \textbf{phone} itunes xbox \textbf{cell}\\   
                          &0.43 & cat cats \textbf{food} vet feed pet \textbf{eat} \textbf{foods}\\
                          \midrule
\multirow{5}{*}{WeTe}  &0.47   & cup add chicken oil cheese salt minutes flour\\
                        &0.47   & god religion believe bible jesus christian religious church\\
                        &0.43   & eat food weight eating diet healthy lose fat\\
                        &0.36    &download ipod video dvd file music player files\\
                         &0.36 & dog dogs cat vet puppy cats animals pet\\
                         \midrule
\multirow{5}{*}{CLNTM}  &0.48   & \textbf{pokemon team diamond pearl earthquake surf beam dragon}\\
                        &0.42   & add minutes oil just cook salt heat garlic\\
                        &0.42   & \textbf{pokemon trade pearl diamond fc code battle shiny}\\
                        &0.41    &laptop pc card memory graphics ram mb processor\\
                        &0.40 & video dvd download format convert videos ipod movie\\
                        
 \midrule
\multirow{5}{*}{ContraTopic}  & 0.59& pentium mhz nvidia ghz intel mb geforce laptop\\
                        &0.58& preheated parmesan browned preheat bake grated mozzarella saute\\
                        &0.57& wcs aeropostale servlet jsp fid asp abercrombie pacsun\\
                        &0.56& hhh khali cena batista umaga wwe vs orton\\
                        &0.52& sugar cup cream butter mix chocolate baking flour\\                      
                           \bottomrule 
\end{tabular}}
  \vspace{-0.15in}

\label{tab:yahoo}

\end{table}

\textbf{Topic Descriptions for 20NG in ContraTopic}:
\begin{itemize}
    \item \textbf{Topic 1: Armenian Genocide and Turkish-Armenian Relations.}
This topic revolves around historical and geopolitical aspects related to the Armenian Genocide, Armenian people, and their interactions with Turkey and Azerbaijan. It highlights the historical conflict and tensions between Armenia and Turkey, as well as the involvement of the Soviet Union and Azerbaijan in the region.
    \item \textbf{Topic 2: Medical Research and Treatment.}
This topic focuses on various aspects of healthcare, including patients, medical conditions, diseases, cancer, drugs, and treatment. It suggests discussions related to medical studies, advancements in treatment methods, and the exploration of different drugs for managing diseases.
    \item \textbf{Topic 3: Christianity and Faith.}
This topic delves into matters of faith, particularly centered around Christianity. It includes references to God, Jesus, the Church, the Bible, and the beliefs and practices of Christians. It reflects discussions about religious faith, the significance of Jesus and Christian teachings, and the concept of heaven.
    \item \textbf{Topic 4: Software Applications and Development.}
This topic centers on software development and related concepts. It mentions servers, applications, clients, user interfaces. It suggests discussions about developing software, creating widgets and applications, exporting data, and user interactions in the context of Unix-based systems.
    \item \textbf{Topic 5: Image Formats and Graphics Software.}
This topic revolves around graphic design, specifically discussing image formats such as JPEG and GIF. It touches upon graphics software and the representation of images using different color formats. It implies discussions about graphic design software, image file types.
\end{itemize}
% \begin{table}[h]
% \centering
% \caption{Generated topics on Yahoo.}
% \scalebox{0.8}{\begin{tabular}{c|l}
% \toprule
% NPMI&Words\\
%  \midrule
% 0.59& pentium mhz nvidia ghz intel mb geforce laptop ram pci\\
% 0.58& preheated parmesan browned preheat bake grated mozzarella saute sprinkle oven\\
% 0.57& wcs aeropostale servlet jsp fid asp abercrombie pacsun catalog stores\\
% 0.56& hhh khali cena batista umaga wwe vs orton undertaker booker\\
% 0.52& sugar cup cream butter mix chocolate baking flour vanilla tsp\\
% 0.51& wr qb fantasy rb johnson john jackson jones reggie te\\
% 0.51& hair frizz frizzy curly bangs straightener perm wavy hairstyles straighten\\
% 0.50& video dvd converter mpeg converters format videos divx wmv youtube\\
% 0.50&  weight body lose fat exercise calories lbs burn gain pounds\\
% 0.50& internet wireless router cable usb connect network connection modem plug\\ 
% \hline 
% \end{tabular}}

% \label{tab:yahoo}
% \end{table}

\textbf{Topic Descriptions for Yahoo in ContraTopic:}
\begin{itemize}
    \item \textbf{Topic 1: Computer Hardware.}
This topic revolves around computer hardware components and specifications. The topic seems to focus on aspects like processors, graphics cards, memory, and connectivity.
    \item \textbf{Topic 2: Cooking Instructions.}
This topic provides instructions and techniques for cooking. The topic suggests actions such as preheating, baking, sautéing, and sprinkling ingredients to create delicious dishes.
    \item \textbf{Topic 3: Fashion Brands and Stores.} This topic relates to fashion brands and retail stores. The topic indicates a focus on brands, online catalogs, and physical stores associated with clothing and fashion.
    \item \textbf{Topic 4: Professional Wrestling.} This topic revolves around professional wrestling and wrestling personalities. The topic suggests discussions or comparisons between various wrestlers, their matches, and their involvement in the WWE (World Wrestling Entertainment).
    \item \textbf{Topic 5: Baking Ingredients.} This topic focuses on ingredients used in baking. The topic suggests the use of these ingredients in recipes to create baked goods.
    % \item \textbf{Topic 6: Fantasy Football Players.} This topic relates to players in fantasy football. It includes terms like "WR" (wide receiver), "QB" (quarterback), "fantasy," "RB" (running back), and various player names like Johnson, John, Jackson, Jones, and Reggie. The topic implies discussions about these players and their performances in fantasy football leagues.
    % \item \textbf{Topic 7: Hair Styling.} This topic focuses on hair styling and treatments. The topic suggests discussions about dealing with frizz, styling curly or wavy hair, using straighteners, and achieving desired hairstyles.
    % \item \textbf{Topic 8: Video Conversion.} This topic relates to video conversion and file formats. The topic suggests discussions or inquiries about converting videos between different formats, including popular formats like MPEG, DivX, and WMV.
    % \item \textbf{Topic 9: Weight Loss and Exercise.} This topic revolves around aspects related to weight, body, and exercise. It likely involves discussions about losing fat, burning calories, gaining or losing pounds, and the importance of exercise in maintaining a healthy lifestyle.    \item \textbf{Topic 10: Internet Connectivity.} This topic revolves around internet connectivity and related devices. The topic implies discussions or instructions on connecting to the internet, using routers, cables, modems, and wireless technology.
\end{itemize}

% \begin{table}[h]
% \centering
% \caption{Generated topics on NYTimes.}

% \scalebox{0.8}{\begin{tabular}{c|l}
% \toprule
% NPMI&Words\\
%  \midrule
% 0.61& palestinian israeli israel arafat yasser west israelis violence gaza peace\\
% 0.55& afghanistan taliban laden bin afghan terrorist qaida war pakistan islamic\\
% 0.52& cup chopped add pepper onion garlic minutes olive cooked tablespoon\\
% 0.46& quarterback game coach bowl touchdown yard offense football defensive nfl\\
% 0.43& gore election ballot voter vote votes bush florida al recount\\
% 0.41& elian cuban castro miami cuba gonzalez fidel juan boy church\\
% 0.40& flight airlines airport air airline plane passenger jet pilot aircraft\\
% 0.40& run inning manager hit yankees sox game pitch pitcher homer\\
% 0.39& movie film character comedy movies hollywood starring actor episode drama\\ 
% 0.36& food eat restaurant cream meal cheese flavor chocolate bread kitchen\\
% \hline
% \end{tabular}}

% \label{tab:nytimes}
% \end{table}

\begin{table}[t]
\centering
\caption{Generated topics on NYTimes.}
\resizebox{\linewidth}{!}{
\begin{tabular}{c|c|l}

\toprule
Models&NPMI&Topic Word Examples\\

 \midrule
\multirow{5}{*}{LDA}    &0.62  &palestinian israel israeli arafat yasser peace sharon israelis\\
                         &0.50  & russian russia soviet vladimir putin moscow union chechnya\\
                         &0.42  &\textbf{con una mas las por como dice los}\\
                          &0.40 & cup minutes add tablespoon oil pepper sugar teaspoon\\
                           &0.35 &military army taliban afghanistan forces war troop soldier\\
                           \midrule
\multirow{5}{*}{ETM}   &0.69   & palestinian israeli israel arafat peace yasser israelis arab\\
                         &0.49  & cup tablespoon add teaspoon sauce minutes sugar butter\\
                        &0.45   & military war taliban forces afghanistan army afghan soldier\\
                         &0.44  & game coach quarterback yard football bowl touchdown defensive\\   
                          &0.41 & film movie character actor movies comedy starring hollywood\\
                          \midrule
\multirow{5}{*}{WeTe}  &0.67   & dicen informacion algunos telefono para notas tienen estan\\
                        &0.51   & habla clase aparecen nuevas pocos busca sigue tiempos\\
                        &0.44   & bush republican campaign bill clinton gore house\\
                        &0.35    &run hit \textbf{game} baseball \textbf{team} \textbf{season} manager league\\
                         &0.35 & \textbf{team} \textbf{game} \textbf{season} lay player win won point\\
                         \midrule
\multirow{5}{*}{CLNTM}  &0.53   & pelzer berkley kiyosaki how zukav lechter bantam jove distinguishable\\
                        &0.49   & kostunica serbian vojislav belgrade serbia yugoslav serb yugoslavia\\
                        &0.46   & additionally toder oder eta column pageex nytsyn rickc\\
                        &0.43    &pga bogey birdie birdies putt fairway par tee\\
                        &0.40 & anos disney sus parte gran ser entre estan\\
                        
 \midrule
\multirow{5}{*}{ContraTopic} & 0.78& economia dedicada notas cubrir transmiten comercio temas expertos\\
& 0.72&erstad spiezio glaus bengie schoeneweis darin disarcina garret\\
& 0.71&palestinian israeli israel arafat israelis yasser sharon jerusalem\\
& 0.61&taliban afghanistan laden afghan bin pakistan islamic osama\\
& 0.56&laker nba neal shaquille bryant kobe phil jackson\\                          
                           \bottomrule 
\end{tabular}}

\vspace{-0.1in}
\label{tab:nytimes}

\end{table}

\textbf{Topic Descriptions for NYTimes in ContraTopic:}
\begin{itemize}
    \item \textbf{Topic 1: Economy and Trade Expertise.} This topic likely focuses on economic issues, trade, and expertise in Spanish. It may involve discussions about economic activities, market trends, trade policies, and insights shared by experts in the field.
    \item \textbf{Topic 2: Baseball Players } This topic revolves around baseball players and possibly specific teams. It includes mentions of players such as Darin Erstad, Scott Spiezio, Troy Glaus, Bengie Molina, Scott Schoeneweis, and Garret Anderson, suggesting discussions about their careers.
    \item \textbf{Topic 3: Israeli-Palestinian Conflict.} This topic centers on the Israeli-Palestinian conflict and related issues. It includes references to Palestinian and Israeli identities, key figures like Yasser Arafat and Ariel Sharon, as well as Jerusalem, the contested city central to the conflict.
    \item \textbf{Topic 4: Taliban and Osama bin Laden.} This topic revolves around the Taliban, Afghanistan, and Osama bin Laden, focusing on Islamic extremism and terrorism. It likely involves discussions about the Taliban's rule in Afghanistan, Bin Laden's involvement in terrorism, and related geopolitical implications.
    \item \textbf{Topic 5: Los Angeles Lakers and NBA.} This topic pertains to the Los Angeles Lakers basketball team and the NBA. It includes references to players such as Shaquille O'Neal and Kobe Bryant, as well as coach Phil Jackson, suggesting discussions about their careers, achievements, and the team's performance in the NBA.

\end{itemize}

% \section{Limitation}
% The limitations of this study can be summarized as follows.
% While our method primarily focuses on topic-wise contrastive learning to enhance topic interpretability, there is room for further improvement in the quality of document representation compared to certain baselines.
% Subsequent research can explore a unified multi-level contrastive learning framework that incorporates both topic-wise and document-wise approaches, aiming to enhance both topic interpretability and document representation.
% Despite efforts to diversify participant recruitment, the human evaluation predominantly involved individuals with high education levels in science-related fields and a strong inclination towards technology.
% % Additionally, the number of participants was limited.
% In ContraTopic, we utilize pre-computed NPMI scores from the training dataset to evaluate topic interpretability to humans, but leveraging the remarkable capabilities of large language models could offer a superior alternative for more human-aligned measurements.

\section{Limitation}
We consider the following limitations also as our future work.
\textbf{For the methodology,} while our method primarily focuses on topic-wise contrastive learning to enhance topic interpretability, there is room for further improvement in the quality of document representation compared to certain baselines.
Subsequent research can explore a unified multi-level contrastive learning framework that incorporates both topic-wise and document-wise approaches, aiming to enhance both topic interpretability and document representation. Besides, we utilize pre-computed NPMI scores from the training dataset to evaluate topic interpretability to humans, but leveraging the remarkable capabilities of large language models could offer a superior alternative for more human-aligned measurements.
\textbf{For the extra computational cost, }though our method does not demand any human-annotation cost, we still need to compute the NPMI matrix on the corpus.
Though the extra computational expenses are acceptable as we discussed in the "Computational Analysis" section, there is still some room for improvement.
\textbf{For the online setting,} our method is designed under the common offline setting of topic modeling. An important future work can be to extend our method to an online setting where documents are partitioned into time slices~\cite{alsumait2008line, lau2012line}.
\textbf{For the experiments, }despite efforts to diversify participant recruitment, the human evaluation predominantly involved subjects with high education levels in science-related fields.

\section{Conclusion}

We propose ContraTopic, an innovative framework for NTMs, with the aim of enhancing the interpretability of generated topics.
ContraTopic integrates human cognition of topic interpretability into the training process by incorporating a topic-wise contrastive regularizer, which serves as a differentiable surrogate for evaluation metrics.
%Our regularizer considers the most relevant words within the same topics as positive samples, while treating words from different topics as negative samples.
Through contrastive learning, our regularizer encourages topic coherence by bringing positive samples closer together and promotes topic diversity by pushing negative samples apart.
Extensive experiments demonstrate that ContraTopic surpasses competing methods in generating highly interpretable topics and deriving document representations.

\section{Acknowledgement}
This work is supported by the National Natural Science Foundation of China (No.82241052).

\bibliographystyle{IEEEtran}
\bibliography{aaai24}

\end{document}